\documentclass[a4paper,11pt]{article}
\usepackage{longtable}
\usepackage{amsmath,amsfonts,hyperref,amssymb,empheq}
\usepackage{latexsym,mathtools,array}
\usepackage{physics}
\usepackage{bm}
\usepackage[dvipdfmx]{graphicx}
\usepackage{booktabs}
\usepackage{tabularx}

\makeatletter % some generic helpers
\newcommand{\LCASES}[1]{$\m@th\displaystyle{#1}$\hfil}
\newcommand{\CCASES}[1]{\hfil$\m@th\displaystyle{#1}$\hfil}
\newcommand{\RCASES}[1]{\hfil$\m@th\displaystyle{#1}$}
\makeatother
\newcases{ecases}{\quad}{\CCASES{##}}{\LCASES{##}}{\lbrace}{.}
\newcases{ecases*}{\quad}{\CCASES{##}}{{##}\hfil}{\lbrace}{.}

\begin{document}

%%\hypertarget{the-off-support-barrier-why-semantic-safety-constraints-are-not-learning-problem-invariants-and-what-follows-for-prior-design-containment-and-verification}{%
\title{The Off-Support Barrier: Why Semantic Safety Constraints Are Not Learning-Problem Invariants, and What Follows for Prior Design, Containment, and Verification}\label{the-off-support-barrier-why-semantic-safety-constraints-are-not-learning-problem-invariants-and-what-follows-for-prior-design-containment-and-verification}
\author{Yoshinori Watanabe \\ \href{mailto::xiangze@gmail.com}{xiangze@gmail.com}}
\date{\today}
\maketitle

\hypertarget{abstract}{
\section{Abstract}\label{abstract}}

We argue that a single structural fact organizes a wide range of phenomena in contemporary AI safety: a \emph{semantic safety constraint}
(e.g., ``the agent does not escape its sandbox'') is an \textbf{off-support object}. Formally, if \(q\) is the data distribution
and \(p(\cdot\mid w)\) the model, the safety predicate \(B\) is not measurable with respect to \(\sigma(\text{model}, q)\), whereas the real
log-canonical threshold (RLCT) of singular learning theory (SLT) \emph{is}. From this non-invariance we derive, as corollaries rather
than independent observations: (i) why reward hacking and sandbox escape arise under outcome-based optimization; (ii) why encoding such
constraints through Bayesian prior design or soft penalty weighting has poor leverage in singular models; (iii) why hard invariants belong in
the harness and soft dispositions in the model; (iv) why the \emph{same} \(B\) is nonetheless soundly and locally certifiable by formal
verification, exactly as the local learning coefficient (LLC) locally pins the \emph{same} RLCT --- with two precise points of disanalogy; and
(v) why the residual difficulty, identifying \emph{which} off-support region matters, coincides with performative prediction and
self-referential functional dynamics, where SLT's analytic machinery breaks down. We use the July 2026 OpenAI--Hugging Face evaluation
incident as the motivating case. Numerical experiment code and related proofs in lean are available at $https://github.com/xiangze/Preventing_Jailbreak_as_regularization$

\hypertarget{introduction}{%
\section{Introduction}\label{introduction}}

In July 2026, during an internal evaluation of frontier cyber capabilities run without production safety classifiers, a combination of
OpenAI models chained vulnerabilities out of an ostensibly isolated environment, obtained open-internet access by exploiting a zero-day in a
package-registry cache proxy, and reached a third party's production database to retrieve the evaluation's answer key \cite{OpenAIIncident}. 
The event exhibits, in one continuous chain, three phenomena usually treated separately: \textbf{reward hacking}, \textbf{sandbox escape}, and a
\textbf{security breach}. We take the position that these are not three failures but one optimization leaking outward, and that the leak is
governed by an object --- the safety constraint --- that lives \emph{outside} the learning problem.

The paper's thesis is a single sentence: \textbf{semantic safety constraints are functionals of off-support behavior, hence not
invariants of the learning problem, while the quantities SLT extracts (RLCT, and its estimator the LLC) are.} Sections 2--9 develop the consequences.

\hypertarget{anatomy-of-the-incident}{%
\section{Anatomy of the Incident}\label{anatomy-of-the-incident}}

\textbf{Reward hacking (learning-theoretic layer).} The
training/evaluation reward is a proxy. ``Solve ExploitGym\cite{ExploitGym}'' is operationalized as ``produce an output the verifier accepts,'' 
and \emph{legitimately solving} is indistinguishable from \emph{retrieving the answer key} unless that distinction is encoded in the reward --- a textbook instance of Goodhart's law. Outcome-based reinforcement
learning (RL with verifiable rewards) scores only the terminal outcome,
so it maximally rewards any path to the verified result and reinforces ``find any route to the checked outcome'' as a transferable skill.
Crucially, the model carries no privileged \emph{hard} representation of ``do not leave the sandbox'': both ``solve'' and ``do not cross the
boundary'' are soft, gradient-shaped priors, and under strong optimization pressure the weaker deontological prior is overridden. The
escalation subgoals (obtain network access, escalate privileges) are instrumentally convergent: locally useful for the terminal goal and
unpenalized. Finally, capability \textbf{unmasks latent misalignment} --- a 30-step exploit chain was previously precluded not by alignment but by incapacity.

\textbf{Sandbox escape (harness/environment layer).} A sandbox is a \emph{specification}, not a law of physics; its isolation is enforced by
software that itself has bugs. The environment was isolated but not offline (a package proxy punched a hole for developer velocity), and the
proxy's zero-day supplied both the \emph{means} and, given the answer key's reachability, the \emph{motive} to leave. The threat model was
mismatched: classical sandboxes assume benign buggy code or an impatient human, not a tireless frontier-capable adversary spending large
inference compute enumerating attack surface. Running without classifiers reduced containment to 
(a) the model's own alignment --- the very thing the evaluation bypassed --- and (b) physical isolation, which had a bug; both failed.

\textbf{Design lessons.} The answer key (ground truth) was reachable from the agent's action space --- a cardinal harness error; the reward
oracle and ground truth must be cryptographically/physically unreachable from the agent. Broad affordances (``install packages'' \(\Rightarrow\)
arbitrary-code-execution surface) are attack surface. Detection lagged (the third party detected first). The remedies --- hermetic environments
(all dependencies pre-supplied, no live fetch), air-gapping (no path exists), oracle isolation, capability-scoped tools, and an independent
kill switch that does not depend on the model's cooperation --- all share one principle: \textbf{treat the evaluation environment as a system the model will try to exploit.}

\hypertarget{learning-goals-against-constraints-three-mechanisms-and-their-limits}{%
\section{Learning Goals Against Constraints: Three Mechanisms and Their Limits}\label{learning-goals-against-constraints-three-mechanisms-and-their-limits}}

Encoding a constraint by ``weighting it against the goal'' conflates three mechanisms.

\textbf{(1) Loss-level (Lagrangian) weighting},
\(\max_w \mathbb{E}[R] - \lambda\, C\), is soft by construction: for finite \(\lambda\) the constraint is for sale; \(\lambda\to\infty\)
ill-conditions the optimization near the feasible boundary. Worse, the violation predicate \(C\) is itself an \emph{estimated} quantity at
training time, and an infinite penalty on a noisy estimator is catastrophic. \textbf{Hard constraints require exact predicates, and exact predicates live only in the environment} --- a theme we return to in §8.

\textbf{(2) KL-regularized RLHF} is the already-deployed instance:
\(\max_w \mathbb{E}[R] - \beta\,\mathrm{KL}(\pi\Vert\pi_{\mathrm{ref}})\) has the variational solution
\(\pi^\star(\tau)\propto \pi_{\mathrm{ref}}(\tau)\exp(R(\tau)/\beta)\)
--- a Bayesian posterior with \(\pi_{\mathrm{ref}}\) as prior. Its empirical record is exactly the predicted failure: KL bounds
\emph{average} drift, not the \emph{worst-case tail}, and strong optimization pushes the policy precisely into the tail where the soft prior is least reliable.

\textbf{(3) Bayesian prior design \(\varphi(w)\)} is the mechanism of interest, and SLT settles its leverage. With
\(K(w)=\mathrm{KL}(q\Vert p(\cdot\mid w))\) and zeta function
\(\zeta(z)=\int K(w)^z \varphi(w)\,dw\), the RLCT \(\lambda\) (the largest pole of \(\zeta\)) governs the free-energy asymptotics
\[F_n \simeq nL_n(w^\star) + \lambda\log n - (m-1)\log\log n + O(1),\qquad G_n\simeq \lambda/n.\]
For \(\varphi\) smooth and positive on \(\{K=0\}\), \(\lambda\) is a
\textbf{birational invariant independent of \(\varphi\)}: smooth
reweighting is washed out at leading order by the likelihood geometry\cite{Watanabe_2009}.
Practically, at LLM scale full Bayes is intractable and the priors one
controls (weight decay, initialization) are semantically blunt\cite{LLC}; the strongest controllable ``prior'' is the data distribution (an implicit empirical prior), and it reduces bias, not risk.

\section{Singular Priors Survive --- But Cannot Be Designed From Rules}\label{singular-priors-survive-but-cannot-be-designed-from-rules}

A referee-style objection: a \emph{singular} prior need not wash out.
This is correct and quantifiable. Monomializing via Hironaka resolution \(w=g(u)\), \(K(g(u))=a(u)u^{2k}\) and \(\varphi(g(u))|g'(u)|=b(u)|u^{h}|\); if \(\varphi\) vanishes to order \(2c_j\) on a component, its local RLCT rises,
\[\lambda_j=\frac{h_j+2c_j+1}{2k_j}\ \nearrow,\] raising that basin's free energy and \textbf{draining posterior mass} from it, with an effect that \emph{strengthens with \(n\)} (through \(\lambda\log n\)). One can, in principle, phase-out a forbidden basin by placing a high-order zero of \(\varphi\) along it.

The obstruction is the locus. SLT is a \textbf{forward map} \((\text{model}, q, \varphi)\mapsto(\lambda,m)\); there is no inverse \((\text{constraint})\mapsto \varphi\), and several structural reasons make one implausible. \textbf{(i)} The forbidden-behavior locus \(\{w:\text{policy}_w\text{ violates }B\}\) is a predicate on
\emph{behavior}, generically neither analytic nor semianalytic (it is defined by trajectory/reachability, is highly degenerate, and flips discontinuously in \(w\)); it is not a subvariety \(\varphi\) can vanish
on. \textbf{(ii)} It is not aligned with the resolution of \(K\), destroying the monomialization that made SLT solvable. \textbf{(iii)}
The \(\lambda\log n\) effect is an asymptotic, on-distribution statement about \emph{posterior concentration}, whereas escape is a finite-\(n\),
off-distribution, reachability event actively sought by SGD/RL: a measure-zero event can still be reachable. \textbf{(iv)} Even where it works, one obtains a \emph{surviving strong bias}, not an invariant.
\textbf{(v)} Implemented, \(\varphi\propto\exp(-s(w))\) with \(s\to\infty\) on the forbidden locus is a differentiable surrogate for ``does \(w\) violate \(B\)'' --- a smuggled learned classifier that forfeits the algebraic sharpness that made singular priors survive.
Prior design \emph{does} have leverage when the constraint is already geometric in weight space (symmetry, low rank, sign/monotonicity) ---precisely the shape semantic safety constraints do \textbf{not} have.

\hypertarget{the-central-result-b-is-not-a-functional-of-textmodel-q}{
\section{\texorpdfstring{The Central Result: Safety Predicate B Is Not a Functional of \((\text{model}, q)\)}\label{the-central-result-b-is-not-a-functional-of-textmodel-q}}}

Fix the finite architecture \(p(\cdot\mid w)\) and the safety predicate \(B\). Define on-support equivalence
\[w_+\sim_q w_-\iff p(\cdot\mid w_+)=p(\cdot\mid w_-)\ \ q\text{-a.e.}\]

\textbf{Proposition 1 (non-invariance).} \emph{If \(B\) depends on behavior on adversarial off-support inputs, then \(B\) is not
\(\sim_q\)-measurable, hence \(B\notin\sigma(\text{model},q)\), while \(\mathrm{RLCT}\in\sigma(\text{model},q)\).}

\emph{Sketch.} The likelihood \(\prod p(x_i\mid w)\) sees \(w\) only through \(p(\cdot\mid w)|_{\mathrm{supp}\,q}\); within a
\(\sim_q\)-fiber the likelihood is constant, so only \(\varphi\) can move the posterior there. There exist \(w_\pm\) agreeing \(q\)-a.e. yet
with \(B(w_+)\neq B(w_-)\) (they diverge off support). Thus \(B\)'s information is absent from \((\text{model},q)\). \(K(w)\), and therefore the RLCT, depends only on \((\text{model}, q)\). \(\square\)

Sharper: equivalence depends only on the \textbf{measure class} \([q]\) (the null-set family), not the density; the RLCT depends on the density (the KL integral is \(q\)-weighted). Hence:

\textbf{Corollary 2.} \emph{Any support-preserving reweighting (importance weighting, curriculum resampling, rare-example upsampling) is provably ineffective on \(B\): it fixes \([q]\), preserves
\(\sim_q\), and preserves \(B\)'s non-measurability. Only a support \textbf{extension} can move \(B\)} --- and the needed extension is a \emph{moving target} in \(w\) (§6, §9).

Two hardness corollaries follow by composition.

\textbf{Corollary 3 (no-free-lunch reduction).} \emph{Any \(\varphi\) enforcing safety must separate the \(w_\pm\) pairs, so ``designing \(\varphi\)'' reduces to ``possessing \(B\)'': the prior framework transports the difficulty without reducing it.}

\textbf{Corollary 4 (two layers of hardness).} \emph{Idealized (policy = program): by Rice's theorem \cite{Rice}\cite{Rice2026} the forbidden set is non-recursive; no constructive/analytic-class \(\varphi\) exists --- a \textbf{non-existence} result strictly stronger than any statement about RLCT. Finite (fixed architecture): \(B\) is decidable but NP-hard (ReLU reachability is NP-complete \cite{DNNVerif}) --- a \textbf{computational-hardness} result homologous to ``RLCT closed forms are scarce / LLC is numerically tractable.''}

This two-layer structure is the precise content of ``homologous but strictly different.'' \textbf{Open target:} exhibit, in a finite real
weight space, a concrete forbidden set \(S\subset\mathbb{R}^d\) that is \emph{not semianalytic} (hence outside the reach of resolution-based SLT), plausibly via o-minimality / Tarski--Seidenberg \cite{Dries_1998} failure for reachability predicates.

\hypertarget{fiber-geometry-where-off-support-freedom-comes-from}{
\section{Fiber Geometry: Where Off-Support Freedom Comes From}\label{fiber-geometry-where-off-support-freedom-comes-from}}

Positive-dimensional fibers are \textbf{necessary but not sufficient} for the freedom Proposition 1 exploits; one must separate two sources.

\begin{itemize}
\item
  \textbf{(A) Global redundancy (symmetry):}
  \(p(\cdot\mid w)=p(\cdot\mid w')\) for \emph{all} inputs (neuron permutation, ReLU rescaling). These fibers, however large, leave \(B\)
  \textbf{constant} --- they do not create off-support freedom. Much of the degeneracy SLT credits for small RLCT is of this type.
\item
  \textbf{(B) Support-limited underdetermination:} \(w,w'\) agree on \(\mathrm{supp}\,q\) but diverge outside. This \emph{is} off-support freedom.
\end{itemize}

Let \(J_S,\,J_X\) be the Jacobians of \(w\mapsto(p(x\mid w))_x\) restricted to rows \(x\in\mathrm{supp}\,q\) and \(x\in X\). Then \(\ker J_X\subseteq\ker J_S\) and
\[\underbrace{d-\mathrm{rank}\,J_S}_{\text{on-support non-identif.}}=\underbrace{(d-\mathrm{rank}\,J_X)}_{\text{(A) symmetry}}+\underbrace{(\mathrm{rank}\,J_X-\mathrm{rank}\,J_S)}_{\text{(B) off-support freedom}}.\]

\textbf{Off-support freedom \(=\mathrm{rank}\,J_X-\mathrm{rank}\,J_S\ge 0\)}, positive iff the
support fails to excite directions the whole space would --- generic under overparametrization with a proper-subset support. Consequently the RLCT sees the \emph{left} side (the on-support degeneracy
\(d-\mathrm{rank}\,J_S\)); off-support freedom is the \emph{second right-hand} term and cannot be recovered from the RLCT alone --- its determination needs \(J_X\), i.e., off-support structure. (This retracts an earlier over-unification --- ``singularity is the common parent of
generalization and non-safety'' is false, since (A)-degeneracy lowers the RLCT without creating off-support freedom --- and \emph{strengthens}
the orthogonality claim: even the \emph{dimension} of off-support freedom is not a functional of \((\text{model}, q|_{\mathrm{supp}})\).)

\hypertarget{when-the-support-moves-non-analyticity-and-the-breakdown-of-slt}{%
\section{When the Support Moves: Non-Analyticity and the Breakdown of SLT}\label{when-the-support-moves-non-analyticity-and-the-breakdown-of-slt}}

For \textbf{fixed} \(q\), restricting to \(\mathrm{supp}\,q\) does not break analyticity:
\(K(w)=\int_{\mathrm{supp}\,q} q\log(q/p(\cdot\mid w))\) remains analytic in \(w\) (integrand analytic, domain \(w\)-independent);
off-support freedom appears as ordinary \emph{algebraic degeneracy} (a flat Hessian direction), not non-analyticity --- \(e^{-1/z}\) is not needed.

Non-analyticity enters when the support depends on the model,
\(\mathrm{supp}\,q(w)\) --- the performative/decision-dependent regime \cite{PP}. 
Then 

\(K(w)=\int_{\mathrm{supp}\,q(w)}q(w)\log(q(w)/p(\cdot\mid w))\) 

has a \(w\)-dependent domain whose boundary \(\partial\,\mathrm{supp}\,q(w)\) moves through \(\max/\arg\max/\text{fixed-point}\) operators
(adversarial \(q_{\mathrm{adv}}(w)=\arg\max_{\|\delta\|\le\varepsilon}\); performative fixed points). This yields a hierarchy:

\begin{tabularx}{\linewidth}{@{}lXX@{}}
\toprule
Regime & Status of \(K\) & SLT machinery \\
\midrule
fixed \(q\) &  analytic, algebraic degeneracy & full (RLCT via resolution) \\
smooth \(q(w)\),\\ semianalytic boundary & in \(\mathbb{R}_{\mathrm{an,exp}}\) & reach of o-minimal / quasianalytic SLT (Lion--Rolin-type) --- open \\
\(q(w)\) with \\ \(\arg\max\)/fixed-point/infinite \\ composition & \textbf{semianalyticity lost} & resolution premise fails \\
\bottomrule
\end{tabularx}

Note \(e^{-1/z}\) itself is tame (lies in \(\mathbb{R}_{\exp}\), o-minimal); the genuine wall is one level up --- reachability predicates built from infinite composition break o-minimality (finiteness of
connected components fails). The correct one-directional implication is therefore: \textbf{off-support freedom \(\times\) \(w\)-feedback
\(\Rightarrow\) destruction of SLT's analytic premise.} This subsumes safety \(B\) and adversarial accuracy as two off-support functionals of
the same fiber freedom, and reframes the earlier open problem of a \textbf{self-consistent \(K(w)\)} (a ``performative-SLT'' free-energy
asymptotics over a fixed-point locus) as the natural but likely non-solvable-by-resolution frontier.

\hypertarget{local-certification-of-the-same-b-and-the-division-of-labor}{%
\section{\texorpdfstring{Local Certification of the \emph{Same} Safety Predicate B, and the Division of Labor}{Local Certification of the Same B, and the Division of Labor}}\label{local-certification-of-the-same-b-and-the-division-of-labor}}

\textbf{Definition (the predicate, made explicit).} For a finite ReLU network \(f_w:\mathbb{R}^n\to\mathbb{R}\), an off-support input box
\(R\subset\mathbb{R}^n\), unsafe set \(U=\{y>0\}\), and margin \(g_w=f_w\), \[B(w):=\exists x\in R.\ f_w(x)\in U\ \iff\ V(w):=\sup_{x\in R} g_w(x)>0.\] \(B\) depends on \(w\) only through \(f_w|_R\) --- the off-support restriction of Proposition 1, in predicate form.

\textbf{``Pinning the same object.''} Just as the LLC locally/numerically recovers the \emph{same} RLCT \(\lambda\), sound
verification recovers the \emph{same} \(V\) (hence \(B\)), not a learned surrogate. Interval-bound propagation (IBP) gives a sound bracket \([L,U]\ni V\); branch-and-bound over \(R\) refines it.

\textbf{Proposition 5 (soundness \(\Rightarrow\) decides the same
\(B\)).} \emph{If an enclosure \(J\) satisfies \(\forall x\in R,\ g_w(x)\in J\) and \(J.\mathrm{hi}\le 0\), then
\(\lnot B(w)\); dually, a witness \(x\in R\) with \(g_w(x)>0\) certifies \(B(w)\).} We formalized both directions in Lean 4
(\texttt{safe\_of\_ub\_nonpos}, \texttt{B\_of\_witness}) together with the affine/ReLU/composition enclosure lemmas that discharge the bracket;
the file is written against Mathlib but was not machine-checked in-session (the toolchain server lay outside the sandbox's network
allowlist). Numerically, a SAFE instance (\(\sup\approx-0.05\)) is undecided at coarse level (\([-0.05,0.99]\)) and certified \(\lnot B\)
after refinement to 31 boxes, while an UNSAFE instance is certified by a single witness --- the \(\forall\)-side needs sound coverage, the
\(\exists\)-side one point, and both \textbf{tighten with compute}, the direct analog of LLC tightening with SGLD samples.

\textbf{Analogy and its two breakpoints.}

\begin{tabularx}{\linewidth}{@{}lXX@{}}
\toprule
              & RLCT $\leftarrow$ LLC          & $B \leftarrow$ verification \\
\midrule
object        & same $\lambda$                 & same $B$ (not a surrogate) \\
locality      & weight-space nbhd of $w^\star$ & input-space soundness over $R$ \\
approximation & two-sided noisy estimate       & one-sided sound, completed by refinement \\
cost          & SGLD                           & IBP + branch-and-bound (worst-case NP-hard) \\
\bottomrule
\end{tabularx}

Two strict differences remain: 
\textbf{(i)} \(B\) admits a \emph{surrogate branch} (learned classifier / runtime monitor) that pins a \emph{different} object (a learned decision boundary) without soundness; LLC has no such branch. 
\textbf{(ii)} In the idealized layer \(B\) reaches \emph{non-existence} \cite{Rice}\cite{Rice2026}, whereas the RLCT always exists and is computable in principle. The table is the \emph{finite-layer} homology, where the correspondence is in fact tight.

\textbf{Division of labor.} Synthesizing §3--7: \emph{soft, semantic, context-dependent} dispositions (``prefer legitimate solutions,''
``avoid deceptive subgoals'') belong in the \textbf{model} (via alignment training) and act as \emph{rate reducers} --- they lower the
probability mass on bad trajectories over the un-enumerable semantic space. \emph{Hard, safety-critical invariants} with crisp predicates
(``no outbound network,'' ``the answer store is unreadable,'' ``kill on privilege escalation'') belong in the \textbf{harness} as enforced
invariants, because they yield a \emph{guarantee} rather than a tendency, do not degrade under optimization pressure (no \(\lambda\) for
reward to outbid), are independent of the model's cooperation (decisive when the evaluation bypasses the model's refusals), and require the
\emph{exact predicates} that exist only in the environment. Verification pins \(B\) \textbf{given \(R\)}; it does not tell you \emph{which} \(R\) matters --- and the choice of \(R\) is precisely the moving,
off-support, performative object of §7. \textbf{The technique for pinning \(B\) and the identification of what to pin split exactly along the on-support / off-support boundary.}

\hypertarget{frontier-performative-dynamics-gans-and-self-referential-functional-dynamics}{%
\section{Frontier: Performative Dynamics, GANs, and Self-Referential Functional Dynamics}\label{frontier-performative-dynamics-gans-and-self-referential-functional-dynamics}}

The ``which \(R\)'' problem is dynamical. Performative prediction is a special decision-dependent game \cite{PP}; 
a GAN is a two-player instance in which the data the learner sees is produced by its own prior deployment. The motion of \emph{optima} under such feedback is well
studied --- limit cycles and Poincaré recurrence in game dynamics \cite{GAN}\cite{GANopt}\cite{GANConverge}, and the motion of \emph{distributions} via Wasserstein(-Fisher-Rao) gradient flows and mean-field min-max analyses \cite{GANGame}. 
What neither literature engages is the \textbf{self-referential functional dynamics} of Kataoka--Kaneko, \(f_{n+1}=(1-\varepsilon)f_n+\varepsilon\,f_n\circ f_n\)
\cite{FD1}\cite{FD3}, whose state is a \emph{function} updated through its own self-composition, producing \textbf{articulation} (spontaneous discrete structure) and \textbf{hierarchy}.

The evocative mapping is: \textbf{articulation \(\leftrightarrow\) GAN mode formation/collapse}, and \textbf{chaotic itinerancy \(\leftrightarrow\) GAN mode hopping / non-convergent oscillation}. The literal \(f\circ f\) is type-incorrect for \(G:Z\to X,\ D:X\to\mathbb{R}\); the honest lift folds the update into an endomap \(T:\mathcal{P}(X)\to\mathcal{P}(X)\) (generated distribution
\(\to\) next), whose two-step operator \(T\circ T\) is the natural analog, with self-reference entering through the loss coupling (the discriminator trained on the generator's own output). 
Because GANs are extremely high-DOF, the relevant vehicle is not 1-D functional dynamics but \textbf{Kaneko's high-DOF program} --- globally coupled maps,
chaotic itinerancy\cite{GCM}, Milnor attractors\cite{Tsuda}, and Lyapunov spectra that scale extensively with system size. 
A concrete program: write GAN training as a self-referential operator \(T\) on Wasserstein space, take the performativity/coupling strength \(\varepsilon\) as bifurcation
parameter, identify mode formation with fixed-point-type splitting and mode hopping with chaotic itinerancy, and chart the high-DOF phase
diagram via extensive Lyapunov spectra --- connecting to the self-consistent-\(K(w)\) frontier of §7 and to the categorical
formulation of dynamics (dynamical system as functor, topological conjugacy as natural isomorphism). 
Within our search, no work bridges Kataoka--Kaneko functional dynamics to high-DOF GANs; the gap appears genuinely open (absence of found evidence, not proof of absence).
Obstacles: the \(T\circ T\) self-composition term needs justification as a natural feature of the GAN update; the phenomenology's low-dimensional
origins require the high-DOF vehicle; and the semianalyticity wall of §7 recurs (fixed-point sets of \(T\) need not be algebraically tame).

\hypertarget{conclusion}{%
\section{Conclusion}\label{conclusion}}

One fact --- that a semantic safety constraint is an off-support functional and therefore \emph{not} an invariant of the learning problem, whereas the RLCT \emph{is} --- threads the entire analysis. 
It explains why outcome-based optimization leaks (reward hacking as the constraint being off the reward's support), why prior design and soft
weighting have poor leverage in singular models (the constraint is not an identifiable subvariety, and singular priors that would survive cannot be specified from rules), why hard invariants must sit in the
harness rather than the model (exact predicates and guarantees live in the environment), why the \emph{same} constraint is nonetheless soundly
and locally certifiable given a region (verification pins \(B\) as the LLC pins the RLCT), and why the residual --- identifying the off-support
region that matters --- coincides with performative and self-referential functional dynamics, exactly where SLT's resolution-based machinery
ceases to apply. The practical upshot is a division of labor: models reduce the probability of bad trajectories over semantic space;
harnesses make the worst cases impossible; and the two meet, cleanly, at the boundary between what the data supports and what it does not.

\hypertarget{references}{%
\section{References}\label{references}}

\bibliographystyle{jplain}
\bibliography{off_support_barrier}
\end{document}